# AI-Powered Mental Health Chatbots in Africa: A Systematic Review and Culturally Adaptive Framework

Matshepo LEBESE[1], Pitso TSIBOLANE[2]
[1]*University of Cape Town, Private Bag X3, Rondebosch, Cape Town, 7701, South Africa*
*Tel: +27 21 650 9111, Email: LBSMAT010@myuct.ac.za*
[2]*University of Cape Town, Private Bag X3, Rondebosch, Cape Town, 7701, South Africa*
*Tel: +27 650 1542, Email: pitso.tsibolane@uct.ac.za*

**Abstract:** Mental health challenges in Africa remain under-addressed due to inadequate infrastructure, stigma, and a chronic shortage of professionals. Artificial Intelligence (AI)-powered chatbots are emerging globally as low-cost, accessible tools that can offer psychological support. This paper presents a systematic review of 52 empirical studies published between 2017 and 2025, critically analysing their cultural, linguistic, and infrastructural relevance to African contexts. The findings demonstrate the potential of AI chatbots to improve accessibility, reduce symptoms of anxiety and depression, and expand psychosocial support, yet reveal limited African-specific adaptation. Most systems remain rooted in Western models and English-language designs, leaving critical gaps in local relevance, inclusivity, and sustainability. Based on the findings, the authors develop a Culturally Adaptive Digital Mental Health (CADMH) framework that integrates African cultural values, multilingual design, mobile-first optimisation, and ethical safeguards. The study highlights opportunities and barriers for integrating AI chatbots into African healthcare, offering guidance for research, practice, and policy.

## 1. Introduction

Mental health has increasingly been recognised as an essential component of public health and socio-economic development, yet in many African countries it remains underfunded and neglected [1]. The African region has the lowest density of mental health professionals globally, with fewer than two workers per 100,000 people, a figure that falls drastically below the global median [2]. This gap is compounded by underinvestment from governments, with the World Health Organization reporting that mental health often receives less than one percent of national health budgets [2]. The impact of this neglect is visible in escalating rates of depression, anxiety, suicide, and psychosocial distress, particularly among young people and vulnerable populations [7]. Stigma continues to limit help-seeking behaviour, while geographic distance, poverty, and weak infrastructure obstruct access to care even for those who are willing to seek it [1][2].

Against this background, digital health technologies have been proposed as supplementary solutions. Artificial Intelligence-powered chatbots have emerged globally as promising interventions, delivering support through natural language conversations on mobile devices. These systems combine natural language processing (NLP), machine learning (ML), and in some cases large language models (LLMs), to engage users in psychological dialogues, provide behavioural interventions, and detect crisis scenarios [3][5][12]. Research in high-income contexts has shown that chatbots such as Woebot, Wysa, and Tess can reduce depressive symptoms, enhance user engagement with therapy, and provide scalable services around the clock [4][13][14]. In contexts where the ratio of clinicians to population is low, such as in Africa, the promise of scalable low-cost support appears attractive.

However, the trajectory of chatbot development has largely followed Western epistemologies and infrastructures [8]. Chatbots have been trained on predominantly Euro-American corpus material, developed for urban populations with stable connectivity, and evaluated within high-resource healthcare systems [9][12]. As a result, their transferability to African contexts is not straightforward. The African continent is marked by extraordinary cultural diversity, multilingualism, spiritual frameworks, and local idioms of distress that do not map neatly onto Western psychiatric categories [15][16]. Furthermore, infrastructural challenges, including unreliable electricity, high data costs, and patchy mobile broadband, raise fundamental questions about the accessibility of AI-driven services [7][10].

The neglect of cultural and infrastructural adaptation risks reproducing digital inequalities [11]. Africa's rich traditions of communal support, ubuntu ethics, and local healing practices provide both opportunities and challenges for digital mental health innovation [17]. Chatbots that are simply translated into African languages without cultural embedding are unlikely to resonate with users or build trust [16]. Systems that fail to operate offline or on low-cost feature phones may exclude large segments of the population [18]. Ethical considerations also weigh heavily: questions of data privacy, consent, and algorithmic bias are particularly acute in low- and middle-income countries where regulatory frameworks remain emergent [19][20].

This paper addresses these gaps by systematically reviewing the global literature on AI-powered mental health chatbots, with a critical focus on their implications for Africa. Accordingly, the following research questions guide this study. The primary research question (PRQ) is: *How do AI-powered mental health chatbots address cultural and contextual needs, and what are the implications for African healthcare systems?* The sub-questions are:

- SRQ1: *What are the current definitions, applications, and technological foundations of AI-powered mental health chatbots?*
- SRQ2: *How do existing chatbots incorporate cultural and linguistic adaptation?*
- SRQ3: *What benefits, limitations, and ethical concerns have been reported?*
- SRQ4: *What gaps exist in addressing African contexts, and how can these inform a culturally adaptive framework?*

### *1.1 Objectives*

The purpose of this research is threefold. First, it seeks to synthesise evidence from the global deployment of AI-powered mental health chatbots, focusing on definitions, applications, and technological foundations [3][5][14]. Second, it assesses the extent to which cultural and linguistic adaptations have been incorporated into the design and implementation, highlighting lessons that may be relevant for Africa [12][15]. Third, it considers reported benefits, limitations, and ethical concerns, identifying critical research gaps and proposing a framework for culturally adaptive deployment on the continent [10][11].

The remainder of this paper is structured as follows. Section 2 outlines the research methodology and review process. Section 3 describes the technological basis of AI chatbots. Section 4 discusses global developments in adaptation. Section 5 presents the results of the review, while Section 6 highlights the business and policy benefits. Section 7 concludes with implications, the CADMH framework, and recommendations for future work.

## 2. Methodology

The study employed a Systematic Literature Review (SLR) methodology to ensure a rigorous, transparent, and reproducible synthesis of knowledge on AI-powered mental health chatbots. This choice aligns with established SLR standards in Information Systems and Health Informatics research [21][22]. The review followed the Preferred Reporting Items for Systematic Reviews and Meta-Analyses (PRISMA 2020) guidelines, which provide an evidence-based structure for planning, screening, and synthesising studies.

### *2.1 Planning and Search Strategy*

The review was planned in four stages: defining the research objectives, identifying relevant databases, designing search strings, and developing inclusion and exclusion criteria. The research questions outlined in Section 1 guided all methodological choices. The review aimed to capture both technological and social dimensions of chatbot implementation to ensure breadth and depth.

The databases Scopus, Web of Science, and PubMed were selected for their interdisciplinary coverage of Information Systems, Computer Science, and Health Sciences. Additional manual searches of key journals (such as JMIR Mental Health, Frontiers in Digital Health, and Information Systems Frontiers) were performed to capture in-press or early-release studies. Search terms were grouped under three conceptual clusters:

- *Technology cluster* ("Artificial Intelligence," "machine learning," "conversational agent," "chatbot");
- *Health cluster* ("mental health," "psychological wellbeing," "therapy," "depression," "anxiety");
- *Context cluster* ("cultural adaptation," "low resource," "Africa," "global South").

These were combined using Boolean operators (AND, OR) and truncated where necessary to capture plural or related terms. The following search string was used in Scopus: ("artificial intelligence" OR "machine learning" OR "conversational agent" OR "chatbot*") AND ("mental health" OR "psychological wellbeing" OR "therap*" OR "depression" OR "anxiety") AND ("cultural adaptation" OR "low resource*" OR "Africa*" OR "global South"). The searches were conducted in October 2025.

### *2.2 Inclusion and Exclusion Criteria*

To maintain methodological transparency, a multi-layered inclusion and exclusion protocol was applied. Studies were included if they met all of the following conditions:

- Published between January 2015 and October 2025**,** ensuring recency and capturing the evolution of AI and LLM technologies;
- Peer-reviewed journal articles or full conference papers (to ensure quality);
- Explicit focus on AI-based or AI-enhanced chatbots employing NLP or ML for mental health support, therapy, or wellbeing;
- Empirical or mixed methods design reporting user data, clinical outcomes, or evaluation results;
- Availability in English.

Studies were excluded if they:

- Reported on non-AI, rule-based or fully scripted bots without learning capabilities.

- Addressed physical health only or general well-being without psychological focus;
- Were review papers, editorials, commentaries, or conceptual essays without empirical data (unless used for contextual triangulation);
- Focused purely on back-end algorithmic performance without user interaction or health relevance.

This approach allowed the review to isolate research directly linked to the intersection of AI, human interaction, and mental health care delivery.

*2.3 Screening and Quality Appraisal*

The screening process was conducted in two phases. After de-duplication and title and abstract screening, papers unrelated to mental health, conversational AI, or end-user evaluation were excluded. The remaining studies underwent full-text screening to ensure alignment with the inclusion criteria. Quality assessment was conducted using an adapted checklist based on Kitchenham's IS review framework and the Joanna Briggs Institute appraisal tools. Each paper was rated on:

1. Clarity of objectives and theoretical grounding,
2. Appropriateness of design and data collection,
3. Validity of findings and transparency of limitations, and
4. Consideration of cultural or contextual factors.

Only studies scoring above a minimum quality threshold were included in the final synthesis, as shown in Figure 1.

*2.4 Data Extraction and Synthesis*

A structured coding protocol was employed to systematically extract data from each included study. Codes captured bibliographic details, chatbot characteristics (name, platform, AI technique), target population, research design, cultural or linguistic adaptation strategies, outcome measures, and ethical considerations.

Both quantitative and qualitative syntheses were performed. Quantitative synthesis involved descriptive statistics on publication trends, study types, geographies, and populations. Qualitative thematic synthesis identified recurring themes related to technological capacity, adaptation, ethical design, and implementation challenges.

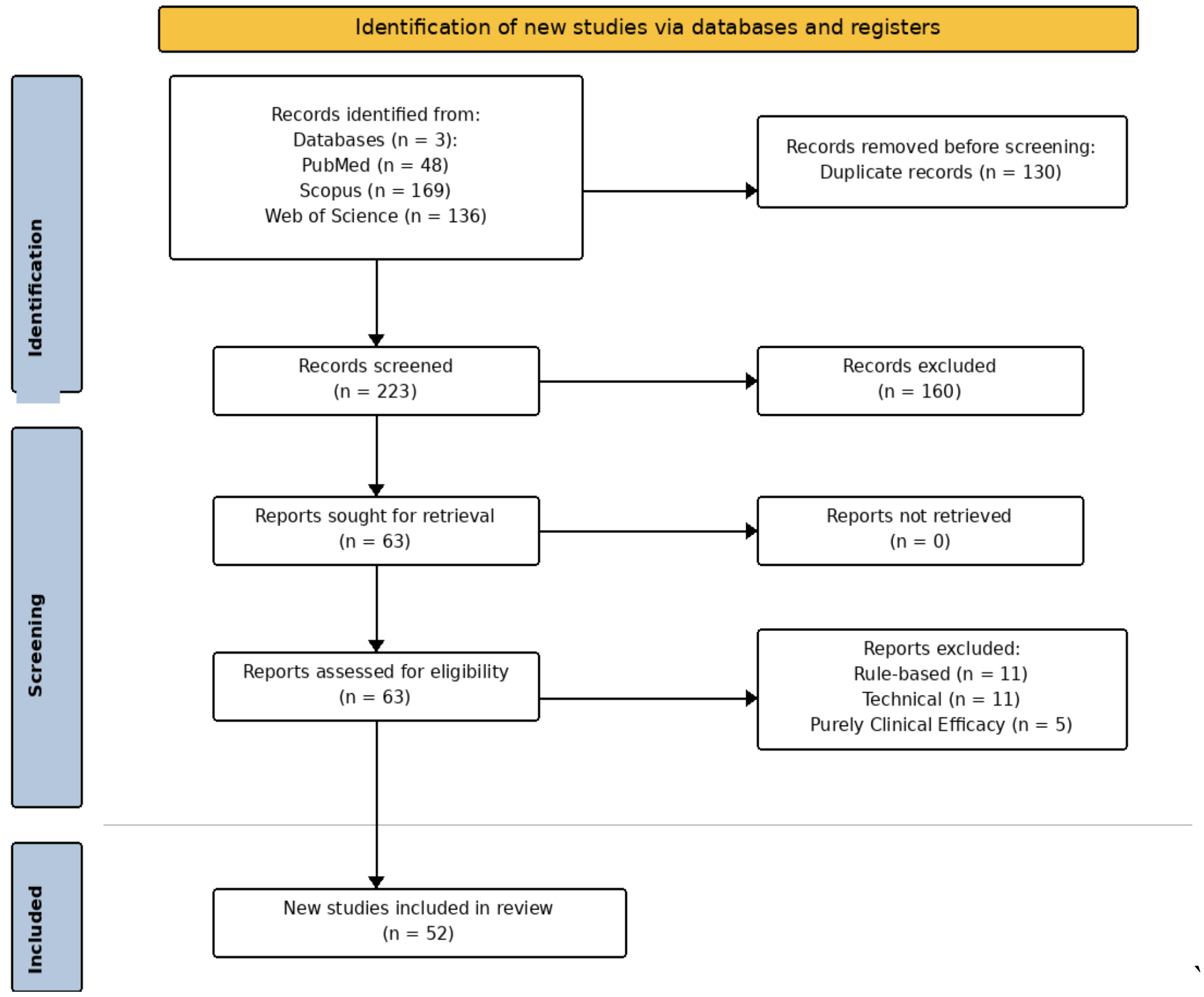

*Figure 1: PRISMA Flow Diagram of Study Selection*

Data extraction captured bibliographic details, technological features, target populations, cultural and linguistic adaptations, outcomes, and limitations. Each study was evaluated for methodological quality and for the extent to which it engaged with cultural and contextual factors. Thematic synthesis was then used to identify recurring patterns, divergences, and gaps. Both quantitative descriptive statistics and qualitative thematic analysis were employed, allowing a comprehensive interpretation of trends [21].

## 3. Technology Description

AI mental health chatbots operate at the intersection of computational linguistics, psychology, and human-computer interaction. They rely on natural language processing to interpret user input, machine learning to personalise responses, and increasingly large language models to sustain empathetic dialogue [5][6][14]. At a technical level, these systems vary from rule-based dialogue engines to advanced generative models. Early chatbots were largely scripted, responding to predetermined keywords or phrases, but more recent systems incorporate transformer architectures capable of learning nuanced conversational patterns [12][19].

The functionalities of these chatbots include psychoeducation, guided self-help modules based on cognitive behavioural therapy, mood and symptom tracking, and conversational support aimed at reducing loneliness or distress [4][13][23]. Some incorporate sentiment analysis to detect crisis states, triggering referrals to emergency services [14][24]. Others integrate with digital platforms, enabling data sharing with clinicians or embedding within health system workflows [25].

Despite these advances, significant limitations persist. Most systems are trained on large datasets in English, reflecting Western psychological discourse [8][9]. This creates both linguistic and cultural bias. A user expressing distress in isiZulu or Kiswahili idioms may not be recognised by the system, resulting in irrelevant or even harmful responses [16]. Similarly, metaphors of suffering rooted in African spiritual traditions may be invisible to models trained on secular Western corpus [17][18].

Technological infrastructure also shapes accessibility. Chatbots that require high bandwidth, or continuous connectivity may be unsuitable in regions with unstable electricity or expensive mobile data [7][10]. Low-end devices common in African settings may not support resource-intensive applications [11][18]. Furthermore, privacy and security architectures designed for contexts with strong regulatory oversight may not translate well into settings where data governance remains weak [19][20].

## 4. Developments

Several global examples illustrate both the promise and the limitations of chatbot adaptation. In Thailand, a chatbot designed for elderly users incorporated Buddhist principles and traditional idioms, which improved resonance and trust [26]. In Peru, interventions for Quechua-speaking adolescents demonstrated the importance of participatory co-design, ensuring that local languages and worldviews were embedded into the chatbot [27]. In Singapore, the mindline.sg platform offered multilingual services in English, Mandarin, Malay, and Tamil, reflecting the multicultural composition of the society and sustaining user engagement across groups [28].

These developments highlight that cultural adaptation is not merely an additional feature but a prerequisite for effectiveness [12][15]. Translation without cultural embedding often results in mechanical and irrelevant interactions, which users quickly abandon [16]. By contrast, participatory design approaches that engage communities in co-creating chatbot dialogues, metaphors, and safeguards produce tools that are more trusted and used [17][27].

## 5. Results

The review of 52 studies revealed several important patterns. First, there is a marked geographic bias: nearly three-quarters of studies were conducted in Asia, Europe, or North America. Only one empirical study examined an African context [7][12] (See Figure 2). This imbalance demonstrates that while the technology is advancing rapidly, its relevance to African populations remains untested.

Second, technological foundations showed variation. Most studies employed natural language processing, with a minority incorporating machine learning for personalisation. Only a small number utilised large language models or multimodal features [14][23]. While technical sophistication is increasing, many studies reported challenges with context comprehension, limited training data, and inadequate recognition of crisis scenarios [19][24].

Third, cultural and linguistic adaptation was limited. Only a quarter of studies explicitly engaged in cultural adaptation, often through participatory design [12][27]. Some offered multilingual support, but most restricted themselves to translation without deeper contextualisation [16][28]. The majority relied exclusively on English [9][12]. The result is a research landscape dominated by Western-centric designs, with little engagement with linguistic or cultural diversity [8][15].

Fourth, the benefits reported across studies included improved accessibility, reduced stigma due to anonymity, high user satisfaction, and symptom reduction for conditions such as anxiety and depression [4][5][13]. Chatbots provided 24/7 availability, allowing users to access support outside traditional service hours. They also offered scalability at relatively low cost, reaching populations that might otherwise be excluded [6][14].

Finally, challenges were frequently reported. These included technical limitations in language processing [16][24], ethical concerns regarding privacy and consent [19][20], low sustained engagement due to repetitive interactions [25], and the inability of chatbots to deal with complex or severe cases [23]. Cultural irrelevance emerged repeatedly as a barrier to user trust and acceptance [12][15].

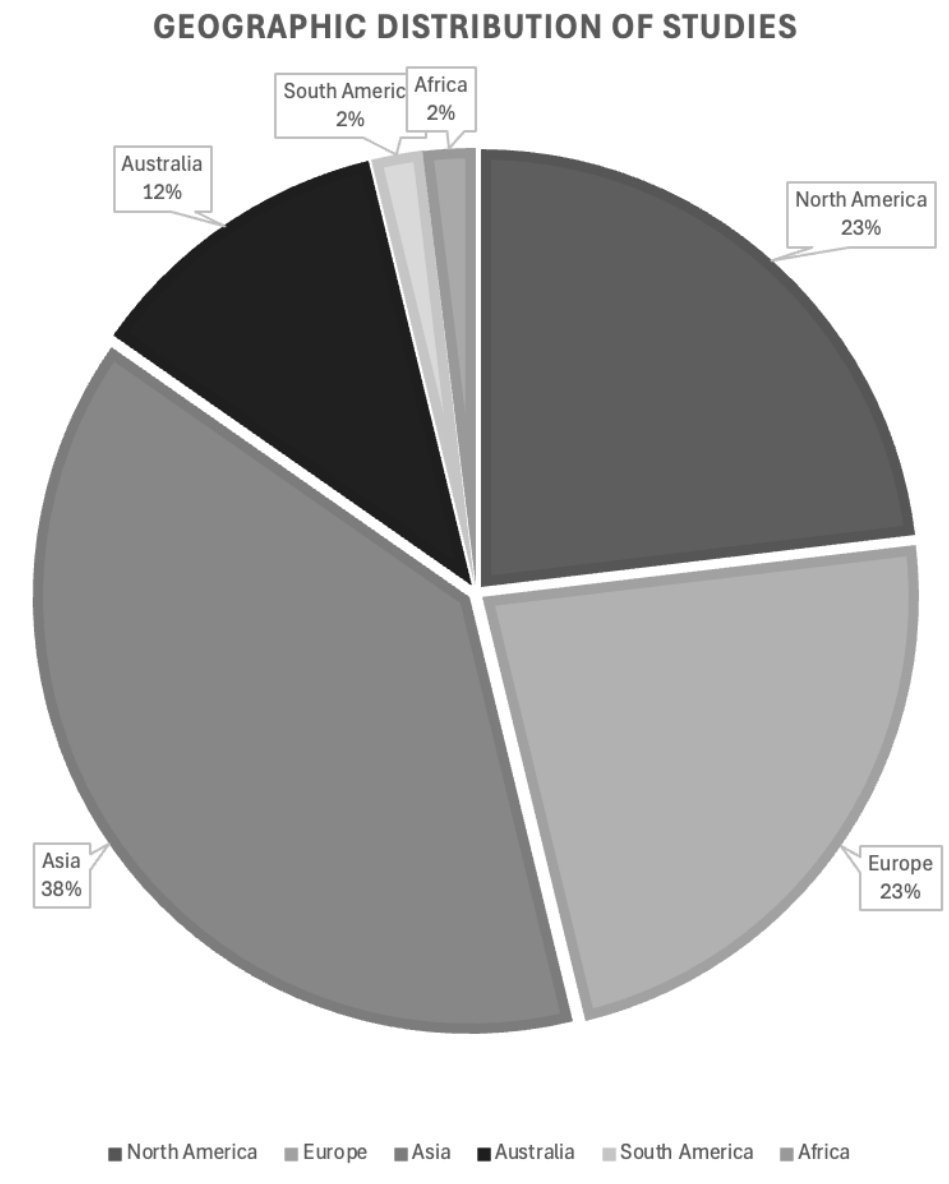

*Figure 2: Geographic Distribution of Studies on AI Mental Health Chatbots*

## 6. Business Benefits

The deployment of AI-powered chatbots in Africa could generate substantial benefits for healthcare systems, governments, and communities. By providing low-cost and scalable psychological support, chatbots could alleviate the severe shortage of mental health professionals [1][2]. They could function as first-line screening tools, triaging users, and referring those in crisis to appropriate human care, thereby reducing clinician workload [13][19].

From a public policy perspective, chatbots align with digital health strategies that seek to expand access and reduce costs [7][11]. They could be integrated into national telehealth platforms, community health worker programmes, or university counselling services [25][28]. For governments with limited resources, chatbots offer an opportunity to extend coverage without proportional increases in expenditure [18].

For communities, chatbots could reduce stigma by offering anonymous and private conversations [4][5]. They could support youth populations who are digitally literate but hesitant to seek formal help [6][14]. They could also be deployed in rural areas, where physical distance from clinics remains a barrier [7][12]. Over time, widespread use of culturally adapted chatbots could normalise conversations about mental health, reducing shame and encouraging help-seeking [15][17].

The economic benefits extend beyond healthcare. By reducing absenteeism, improving productivity, and supporting educational attainment, improved mental health can contribute to broader development outcomes [1][2]. However, realising these benefits depends on careful attention to cultural adaptation, technological accessibility, and ethical safeguards [10][11][19]. Without these, chatbots risk excluding the very populations they are meant to serve.

## 7. The Culturally Adaptive Digital Mental Health (CADMH) framework

In addressing the primary research question, the study shows that AI chatbots hold promise for expanding access to care but that their current designs do not adequately reflect African cultural, linguistic, or infrastructural contexts. Instead, they reproduce global North assumptions, which limits their relevance on the continent.

With regard to the sub-questions, the study has shown, first, that existing definitions and applications of AI chatbots emphasise cognitive behavioural therapy, psychoeducation, and mood tracking, supported primarily by natural language processing and machine learning (RQ1). Second, cultural and linguistic adaptation is minimal, with only a minority of studies pursuing participatory design or multilingual engagement (RQ2). Third, benefits include accessibility, stigma reduction, and scalability, while challenges centre on weak contextual comprehension, ethical concerns, and engagement fatigue (RQ3). Finally, the analysis identifies significant gaps in African contexts, particularly the absence of systems trained on African languages, tested in low-resource conditions, or integrated into local health systems (RQ4).

To address these gaps, the authors propose the Culturally Adaptive Digital Mental Health (CADMH) framework (See Figure 3), which provides design principles for aligning AI chatbots with African realities. This framework emphasises cultural sensitivity, linguistic inclusivity, mobile-first deployment, healthcare integration, and ethical safeguards.

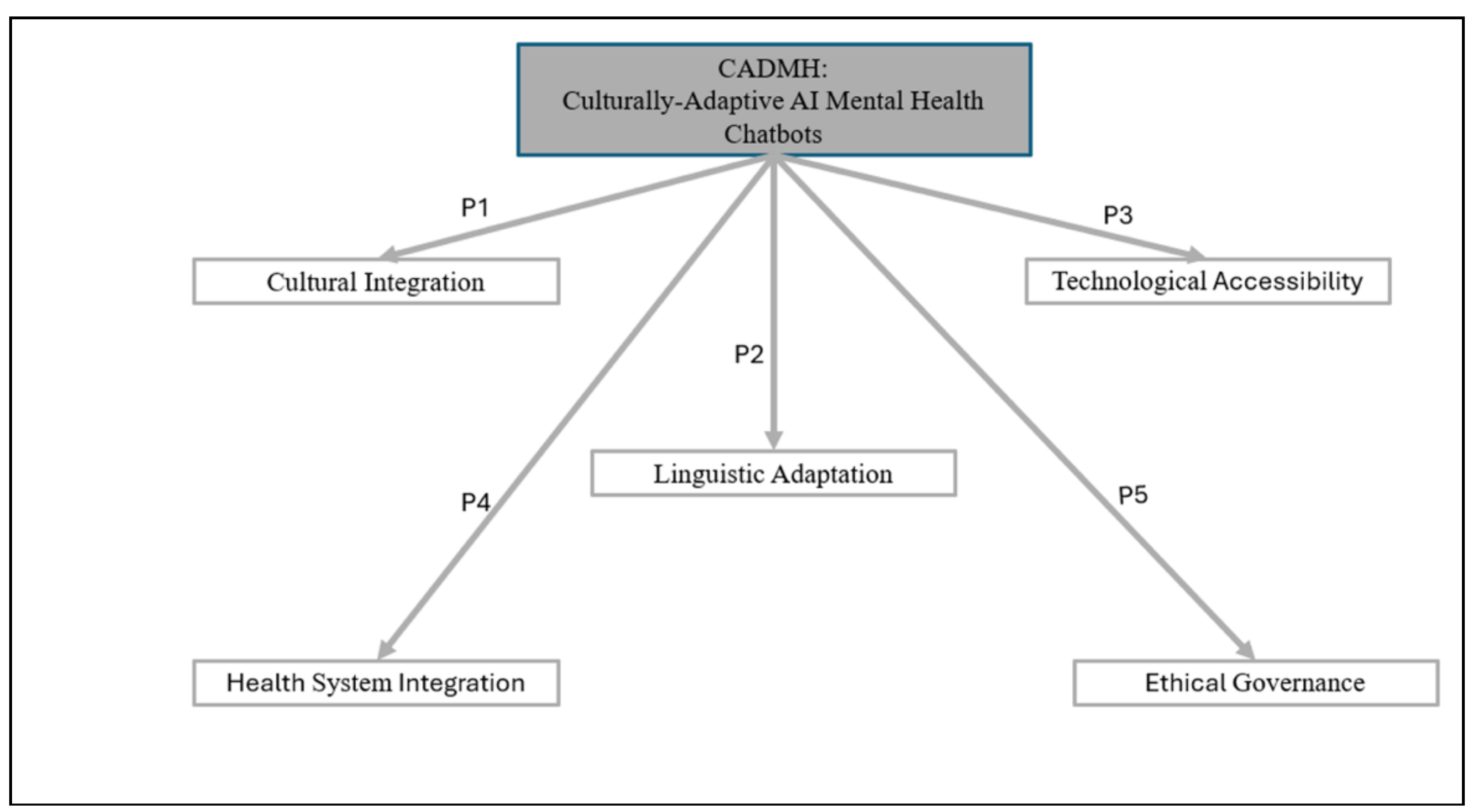

*Figure 3: Culturally Adaptive Digital Mental Health (CADMH) Framework for Africa. P1 = Cultural Adaptation and Engagement; P2 = Linguistic Diversity and Comprehension; P3 = Technological Fit and Accessibility; P4 = Integration and Continuity of Care; P5 = Ethical Governance and User Trust.*

The implications for research and policy are significant. Future work must prioritise the development of context-appropriate tools, evaluated longitudinally in African settings, and embedded in health systems through partnerships that ensure both accountability and sustainability. Policymakers need to strengthen governance frameworks to protect users while enabling innovation.

In conclusion, this study has answered its research questions by demonstrating both the potential and the limitations of AI-powered mental health chatbots for Africa. It makes clear that without deliberate cultural and contextual adaptation, these technologies risk entrenching inequities, but with such adaptation, they can become meaningful tools for addressing Africa's mental health crisis.

### *7.1 Theoretical Propositions of the CADMH Framework*

The Culturally Adaptive Digital Mental Health (CADMH) Framework (Figure 3) was derived from the thematic synthesis of 52 empirical studies reviewed in this paper, interpreted through the lens of socio-technical systems theory and Activity Theory [10][17][19]. It integrates empirical patterns and theoretical constructs to explain how technological, cultural, linguistic, and ethical factors interact in shaping the success of AI-powered mental-health interventions in African environments.

#### *7.1.1 Derivation Process*

The derivation process for the framework unfolded in three analytical stages. First, a conceptual mapping exercise identified recurring variables across the literature, including localisation, language support, trust, infrastructure, and ethics. Second, a cross-domain synthesis linked these elements to the components of an activity system; subjects (users), tools (chatbots), community (health systems), rules (ethical and cultural norms), and outcomes (mental-health improvement). Finally, the elements were organised into five interrelated layers: a) Cultural, b) Linguistic, c) Technological, d) Healthcare, and e) Ethical, each representing a set of design requirements that must co-evolve for successful adoption.

This stepwise derivation process ensured that the framework is both empirically grounded and theoretically coherent. It situates AI chatbots as mediating artefacts within African social systems rather than as isolated technologies.

#### *7.1.2 Core Framework Propositions*

From this analysis, five propositions emerge that may guide future empirical research and policy development:

**Proposition 1 - Cultural Adaptation and Engagement (P1):**

AI-powered chatbots that incorporate indigenous cultural values, collective idioms, and relational principles such as *ubuntu* will generate higher levels of engagement, empathy, and trust than culturally neutral systems.

**Proposition 2 - Linguistic Diversity and Comprehension (P2):**

Chatbots underpinned by multilingual NLP models trained on major African languages and dialects will exhibit superior comprehension accuracy and therapeutic resonance compared with English-only systems.

**Proposition 3 - Technological Fit and Accessibility (P3):**

Solutions optimised for mobile-first, low-bandwidth, and offline operation will achieve greater accessibility and retention among users in low-resource settings than high-bandwidth or web-dependent models.

**Proposition 4 - Integration and Continuity of Care (P4):**

When embedded within primary-care structures, community-health-worker networks, or telemedicine platforms, chatbots can strengthen triage accuracy, continuity of care, and clinician efficiency.

**Proposition 5 - Ethical Governance and User Trust (P5):**

Transparent data governance, culturally appropriate consent, and compliance with local privacy regulations such as POPIA will enhance user confidence and long-term adoption. These propositions can be formalised as hypotheses for quantitative validation or as guiding statements for qualitative exploration, enabling the framework to evolve through iterative empirical testing.

*Pathway to Implementation*

Translating the CADMH framework into practice requires a phased implementation strategy. In the first phase, pilot chatbot deployments should be co-designed with local communities, incorporating multilingual content and culturally grounded dialogue models informed by Propositions 1 and 2. These pilots could target university counselling services or community health worker programmes, where infrastructure and user populations are relatively accessible [25][28]. In the second phase, partnerships with public health institutions, non-governmental organisations, and technology providers should be established to enable iterative evaluation, secure regulatory alignment with frameworks such as the Protection of Personal Information Act (POPIA), and refine chatbot performance using real-world user data. The third phase involves scaling successful pilots through integration with national telehealth platforms and mobile network operators, leveraging low-bandwidth, offline-capable architectures aligned with Proposition 3. Throughout all phases, sustained stakeholder engagement, transparent governance structures, and dedicated funding mechanisms are essential to ensure that chatbots move from proof-of-concept to sustainable, equitable deployment across African healthcare systems.

## 8. Conclusions and Future Research

This paper has presented a systematic review of AI-powered mental health chatbots, highlighting their global progress and their implications for Africa. The review confirms that chatbots can reduce symptoms, improve accessibility, and provide scalable support [3][4][5]. Yet it also demonstrates that the field remains dominated by Western contexts, with little engagement with Africa's cultural and infrastructural realities [7][8][12].

The propositions convert the CADMH framework from a descriptive model into a research programme that articulates testable relationships among cultural adaptation, linguistic inclusion, technology design, and ethical practice. They encourage mixed-methods inquiry, combining quantitative experiments to measure engagement and outcome differences with participatory design studies to refine cultural interfaces.

By foregrounding Africa's linguistic and cultural complexity, the framework extends existing digital-health theories and provides a conceptual basis for comparative studies across regions. It also invites collaboration among computer scientists, psychologists, and public health practitioners to develop shared evaluation metrics.

In essence, the CADMH framework and its propositions offer both an explanatory lens and a roadmap for future innovation. They underline that the effectiveness of AI in mental health will depend not only on algorithmic sophistication but also on cultural resonance, infrastructural realism, and ethical stewardship.

**Acknowledgements**

***Declaration of use of content generated by Artificial Intelligence (AI) (including but not limited to Generative-AI) in the paper.***

The authors acknowledge the use of AI-assisted editorial support for restructuring and compressing earlier drafts into the IST-Africa format. All substantive content is derived from the authors' systematic review and critical analysis.

*Due to space limitations, the list of 52 articles that are part of the SLR are available upon request*